\documentclass[letterpaper, 10 pt, conference]{ieeeconf}
\IEEEoverridecommandlockouts    %
\usepackage{graphics}           
\usepackage{times}              
\usepackage{amsmath}            
\usepackage{amssymb}            
\usepackage{graphicx}
\usepackage{algorithm}
\usepackage[noend]{algpseudocode}
\usepackage{booktabs}
\usepackage{color}
\definecolor{instructioncolor}{rgb}{.5,.5,.5}

\usepackage[font=small]{caption}

\def\secref#1{Sec.~\ref{#1}}
\def\figref#1{Fig.~\ref{#1}}
\def\tabref#1{Tab.~\ref{#1}}
\def\eqref#1{Eq.~(\ref{#1})}

\makeatletter
\usepackage{xspace}
\DeclareRobustCommand\onedot{\futurelet\@let@token\@onedot}
\def\@onedot{\ifx\@let@token.\else.\null\fi\xspace}

\def\etal{{et al}\onedot}
\makeatother

\def\etalcite#1{\etal~\cite{#1}}

\usepackage{array}
\newcolumntype{L}[1]{>{\raggedright\let\newline\\\arraybackslash\hspace{0pt}}m{#1}}
\newcolumntype{C}[1]{>{\centering\let\newline\\\arraybackslash\hspace{0pt}}m{#1}}
\newcolumntype{R}[1]{>{\raggedleft\let\newline\\\arraybackslash\hspace{0pt}}m{#1}}

\renewcommand{\b}[1]{\mbox{\boldmath$#1$}}

\newcommand{\bm}{\b m}

\usepackage{todonotes}

\usepackage{bbold} %
\usepackage[per-mode=symbol, binary-units]{siunitx}
\usepackage{subcaption}
\usepackage{multirow}
\usepackage{arydshln}
\usepackage{makecell}
\usepackage{balance}
\usepackage{cite}
\usepackage{bm}
\usepackage{pifont}

\newcommand{\cmark}{\ding{51}}%
\newcommand{\xmark}{\ding{55}}%

\makeatletter
\let\NAT@parse\undefined
\makeatother
\usepackage[hidelinks]{hyperref}

\title{\LARGE \bf Online Tree Reconstruction and Forest Inventory\\ on a Mobile~Robotic~System}

\author{Leonard Frei{\ss}muth$^{1,2}$ \;\;\; Matias Mattamala$^{1}$ \;\;\; Nived Chebrolu$^{1}$ \;\;\; Simon Schaefer$^{2}$ \\ Stefan Leutenegger$^{2}$ \;\;\; Maurice Fallon$^{1}$%
\thanks{$^{1}$ The authors are with the University of Oxford, UK. \texttt{\{matias, nived, mfallon\}@robots.ox.ac.uk }.}
\thanks{$^{2}$ The authors are with the Technical University of Munich, Germany. \texttt{\{stefan.leutenegger, simon.k.schaefer, l.freissmuth\}@tum.de}}
\thanks{$^{3}$ Our video attachment can be found at \texttt{\url{https://youtu.be/5AJwPEV1ZMU}}.}}

\let\oldtwocolumn\twocolumn
\renewcommand\twocolumn[1][]{%
    \oldtwocolumn[{#1}{
    \begin{center}
          \vspace{-5pt}
        \includegraphics[width=1\textwidth]{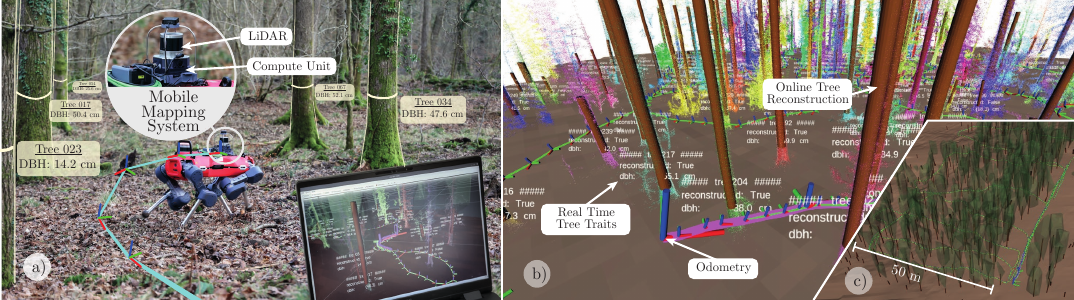}
        \captionof{figure}{The online pipeline for real-time reconstruction of trees running on a mobile robot walking through a forest (a). While acquiring data, a forester can evaluate the mapping process in real-time to evaluate coverage and reconstruction quality. The pipeline is able to reconstruct important tree traits online (b) and has been tested on plots as large as \SI{0.7}{\hectare} (c), which we visualize in our video attachment $^3$.  
	      }
        \label{fig:motivation}
    \end{center}
}] }

\begin{document}
\maketitle
\thispagestyle{empty} 
\pagestyle{empty}

\begin{abstract}
  Terrestrial laser scanning (TLS) is the standard technique used to create accurate point clouds for digital forest inventories.
  However, the measurement process is demanding, requiring up to two days per hectare for data collection, significant data storage, as well as resource-heavy post-processing of 3D data.
  In this work, we present a real-time mapping and analysis system that enables online generation of forest inventories using mobile laser scanners that can be mounted e.g. on mobile robots. 
  Given incrementally created and locally accurate submaps---\emph{data payloads}---our approach extracts tree candidates using a custom, Voronoi-inspired clustering algorithm.
  Tree candidates are reconstructed using an algorithm based on the Hough transform, which enables robust modeling of the tree stem.
  Further, we explicitly incorporate the incremental nature of the data collection by consistently updating the database using a pose graph LiDAR SLAM system.
  This enables us to refine our estimates of the tree traits if an area is revisited later during a mission.
  We demonstrate competitive accuracy to TLS or manual measurements using laser scanners that we mounted on backpacks or mobile robots operating in conifer, broad-leaf and mixed forests.
  Our results achieve RMSE of \SI{1.93}{\centi\meter}, a bias of \SI{0.65}{\centi\meter} and a standard deviation of \SI{1.81}{\centi\meter} (averaged across these sequences)---with no post-processing required after the mission is complete.
\end{abstract}

\section{Introduction}
\label{sec:intro}
In traditional clear-cut forestry, plots of several hectares are felled at once when deemed ready for harvesting or thinning~\cite{matthews1991silvicultural}.
Modern forestry methods aim to minimize the impact on the forest ecosystem by carefully selecting which trees to cut---usually those fully grown or inhibiting the growth of other trees---which is called \emph{continous coverage forestry}~\cite{mason2003frara}.
To support this approach and to assess its impact on the ecosystem requires the systematic data collection of tree locations and relevant tree traits, \emph{forest inventories}, which enable the construction of high-fidelity digital twins of the forest, known as \emph{marteloscopes}.
For producing these marteloscopes, foresters and ecologists traditionally measure the tree traits manually, which is time-consuming and limits the set of tree traits that can be measured.

In order to address these challenges, methods based on terrestrial laser scanning (TLS) enabled a systematic and accurate acquisition of forestry data.
However, they need periodic repositioning of the laser device through the forest, which increases data acquisition time and requires resource-heavy post-processing to determine the forest traits. In addition, vast amounts of raw data are generated, which reach up to \SI{20}{\giga\byte\per\hectare}.
Alternatively, mobile laser scanners (MLS) have a much lower acquisition time as the sensor continuously moves, e.g. on a mobile robot.
MLS methods, however, incur a compromise in sensor accuracy---a TLS sensor achieves millimeter measurement accuracy while MLS scanners are in the centimeter range. As a result, MLS mapping systems suffer from drift in pose estimation.

To address the issues of file size and post-processing time, we propose to amortize the post-processing time by performing the reconstruction online, during data acquisition.
This approach provides immediate feedback on the scanning quality and coverage, and produces a marteloscope immediately after the session ends.
In our initial approach, Proudman \etalcite{proudman2022ras} attempted to achieve this by detecting trees at the sensor rate of the LiDAR and then fusing and correcting the estimates upon loop closure. This required several simplifications in the tree detection and modeling to cope with the sensor frequency, consequently producing inferior reconstructions. This---in combination with the sparsity of single scans---did not allow for faithful estimation of tree parameters.

In this work, we address these limitations and introduce a real-time system to create online forest marteloscopes on mobile robotic platforms.
We use a combination of a pose graph SLAM system and a custom \emph{Tree Manager} module to segment trees, associate measurements over time, and maintain global consistency.
Once there are enough measurements available for a tree, it is reconstructed employing a filtering procedure based on the Hough Transform and a model-averaging reconstruction algorithm.
Although running with limited compute resources, our pipeline produces faithful estimates of important tree traits faster than point clouds are acquired.
Thus, our approach can be considered the first algorithm that enables real-time forest inventory with  reconstructions available as soon as the measurement session has ended.

In summary, we present four contributions in our work:
Our approach is able to
(i) extract relevant tree traits online with accuracy competitive to state-of-the-art post-processing approaches,
(ii) produce a globally consistent tree map of the forest which is built incrementally and updated in real-time,
(iii) robustly detect and fit stacks of oblique cone frustums in the presence of heavy noise, which has been tested on datasets of different tree compositions,
(iv) run on a mobile system - either a quadruped robot or a human-carried backpack.
We have extensively tested our approach using forest data from the UK, Switzerland, and Finland.

\section{Related Work}
\label{sec:related}
\subsubsection{Terrestrial Laser Scanning}
The ecology and forestry community has developed a mature body of literature describing tools for building 
forest inventories using TLS \cite{liang2016jprs, liang2018jprs}.
Methods building high-fidelity models usually employ highly engineered pipelines based on cover sets, clustering, sorting, and cylinder fitting algorithms to reconstruct the tree components \cite{raumonen2013rs, krisanski2021rs}.
While these methods aim to extract maximum information from point clouds, they often incur substantial computational costs, with processing times up to half an hour per tree \cite{raumonen2013rs}.
 
Other methods focus on building a lower fidelity model of the tree that only considers the stem.
These approaches typically begin with Voronoi clustering of the trees \cite{prendes2021gisrs}, followed by individual stem reconstruction~\cite{cabo2018ijaeog, prendes2021gisrs}.
Methods employing clustering of cover sets can reconstruct multiple trees from the point cloud directly \cite{pitkanen2019jprs}.

Stem models usually consist of stacked oblique cone frustums \cite{cabo2018ijaeog, prendes2021gisrs,liang2014tgrs} or a cubic spline \cite{pitkanen2019jprs} that interpolates the diameters, enabling the tree diameter to be interpolated and extrapolated along its height.
These methods involve a reconstruction of a stack of circles. An effective approach has been to filter outliers that are not in keeping with three lower circles \cite{liang2014tgrs, pitkanen2019jprs}.
When coupled with terrain detection, these methods offer an automated procedure to estimate the diameter at breast height (DBH) and other tree traits, such as the total merchantable volume \cite{pitkanen2019jprs, prendes2021gisrs} and stem curvature \cite{prendes2022forests}.
They achieve a Root Mean Square Error (RMSE) for the DBH estimate as low as \SI{7.3}{\milli\meter}~\cite{pitkanen2019jprs}.

While these approaches offer high-fidelity representations and accurate reconstructions, their computational cost makes them unsuitable for online processing.
So instead, in this paper, our emphasis lies on Mobile Laser Scanning, enabling convenient data generation and evaluation by moving the sensor through the plot---in a fraction of the time.

\subsubsection{Mobile Laser Scanning} Other researchers also focus on improving MLS systems, deploying them on backpacks \cite{proudman2022ras, hyyppa2020jprs} or drones \cite{hyyppa2020jprs_drones}.
A notable advantage of MLS systems is that they achieve better coverage of the trees by continuously scanning from multiple perspectives \cite{bienert2018forests}.
The primary challenge facing MLS pipelines, however, is the complex alignment procedure of point clouds.
While short-term missions may suffice with simple integration of odometry for alignment \cite{li2020ijrs}, longer missions need a SLAM system to correct sensor drift in the odometry \cite{hyyppa2020jprs, hyyppa2020jprs_drones, bienert2018forests}.
As we design our pipeline to work with long-duration scans, we also employ a SLAM system to ensure global consistency.
Once a globally consistent map is obtained, MLS pipelines employ various methods to generate a digital terrain model (DTM) to standardize the clouds by height.
This is followed by clustering using learned approaches \cite{li2020ijrs}, the Watershed Algorithm \cite{hyyppa2020jprs_drones}, the QuickShift++ Algorithm \cite{malladi2024icra}, or euclidean clustering \cite{proudman2022ras}.

For modeling a tree, MLS methods usually represent the stem as a single cylinder \cite{proudman2022ras}, a stack of oblique cone frustums \cite{bienert2018forests, liu2021ieeeaccess, hyyppa2020jprs_drones}, or polynomial curves \cite{hyyppa2020jprs, hyyppa2020jprs_drones} fitted to a stack of circles.
The most promising results are achieved by curve models with RMSE for the DBH of \SI{0.6}{\centi\meter}. Non-curve-based methods achieve an RMSE for the DBH as low as \SI{1.14}{\centi\meter} \cite{prendes2021gisrs}.
All these methods process point clouds in post-processing after acquiring all data and leaving the forest. 
To support foresters in gathering high-quality data and reduce reconstruction time, our focus is on an online approach that provides a real-time visualization as the map and reconstructions are being generated.

It is hard to compare the algorithms of related work to our datasets as the complex pipelines and datasets are usually not accessible or applicable to our data.
Instead, we compiled the reconstruction results of state-of-the-art approaches in \tabref{tab:comparison_sota} to assess the performance of our pipeline. 
As can be seen, our approach achieves competitive results while at the same time being able to run online, which is novel to the field.

\begin{table}[]
  \centering
  \begin{tabular}{l|c|c|c}
                                            & \begin{tabular}[c]{@{}c@{}}RMSE \\ DBH $\downarrow$ \end{tabular} & \begin{tabular}[c]{@{}c@{}}Plot \\ Types \end{tabular} & \begin{tabular}[c]{@{}c@{}}online\\ capability\end{tabular} \\ \hline\hline
  Meher et al. \cite{malladi2024icra}       & \SI{11.8}{\centi\meter}                                           & C                                                      & \xmark                                                         \\
  Bienert et al. \cite{bienert2018forests}  & \SI{3.8}{\centi\meter}                                            & M, D                                                   & \xmark                                                         \\
  Liu et al. \cite{liu2021ieeeaccess}       & \SI{2.0}{\centi\meter}                                            & M                                                      & \xmark                                                         \\
  Bauwens et al. \cite{bauwens2016forests}  & \SI{1.1}{\centi\meter}                                            & C, M, D                                                & \xmark                                                         \\
  Hyppä et al. \cite{hyyppa2020jprs_drones} & \textbf{\SI{0.6}{\centi\meter}}                                   & C                                                      & \xmark                                                         \\
  Ours                                      & \SI{1.9}{\centi\meter}                                            & C, M, D                                                & \cmark                                                             
  \end{tabular}
  \caption{Comparison of state-of-the-art approaches for building forest inventories from MLS point clouds. The RMSE for the DBH is given for each approach. The plot types are C: Coniferous, M:~Mixed, D: Deciduous.}
  \label{tab:comparison_sota}
\end{table}

\section{Method}
\label{sec:main}

\begin{figure}[htbp]
  \centering
  \begin{minipage}{\linewidth}
    \centering
    \includegraphics[width=0.95\linewidth]{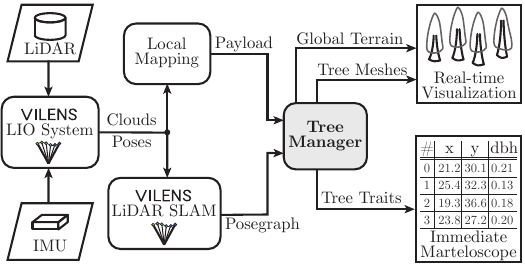} 
    \caption{Overview of our proposed online tree reconstruction pipeline. The central \textit{Tree Manager} is fed with payload clouds from the local mapping module as well as the pose graph from the LiDAR SLAM system. It generates a real-time visualization and constructs a marteloscope.}
    \label{fig:pipeline}
  \end{minipage}
  \begin{minipage}{\linewidth}
    \centering
    \includegraphics[width=0.95\linewidth]{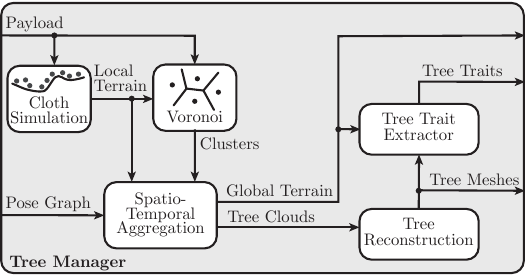}
    \caption{Overview of the Tree Manager module. Terrain models are produced by a cloth simulation filter (CSF) and clusters are computed by a Voronoi-inspired algorithm. We aggregate them over time in a globally consistent manner using the SLAM pose graph. After reconstruction of the trees, we can extract important tree traits.} 
    \label{fig:tree_manager}
  \end{minipage}
\end{figure}

The pipeline we propose, as shown in (\figref{fig:pipeline}), builds on top of our pose graph SLAM system, which is fed by a LiDAR inertial odometry (LIO) module~\cite{wisth2019ral}.
The \emph{local mapping} module integrates point clouds to local submaps---data \emph{payloads}---to increase density.
A constant stream of payloads is the input to our \emph{Tree Manager} (\figref{fig:tree_manager}).
The Tree Manager builds a \emph{Local Terrain Model} and \emph{segments} the trees. It uses the SLAM pose graph to aggregate measurements for tree instances in a \emph{spatio-temporal} manner.
With enough data, the tree is \emph{reconstructed} and \emph{tree traits} are extracted.
The result of the pipeline is an online visualization of the map and a marteloscope immediately after data acquisition.

In our work, we use two coordinate frames:
The map frame $\mathtt{M}$ for final reconstruction and the moving sensor frame $\mathtt{S}^t$ at timestamp $t$, for representing raw measurements.

\subsection{Local Mapping}
Using the poses provided by the odometry and by integrating loop closures, VILENS builds a globally consistent and time-varying pose graph. This graph comprises stamped transformations $^\mathtt{M}\mathbf{T}_{\mathtt{S}^t}$ from sensor frame to map frame.

The LIO system outputs a point cloud at a rate of 10 Hz.
Although the density of an individual scan is sufficient for odometry, it is not dense enough for faithful tree reconstruction. 
This is why we integrate individual measurements along a trajectory of \SI{20}{\meter} using the poses provided by the LIO, which we call a \emph{payload} (abbreviated $\text{pl}$).
We represent the payload in $\mathtt{S}^{t}$ with $t$ referring to the center timestamp of the \SI{20}{\meter} long trajectory.
Using $t$ as a unique identifier, the payload cloud is attached to the SLAM pose graph.
To reduce computational load, we first downsample the cloud using a voxel filter with a resolution of \SI{1}{\centi\meter}.
Secondly, we remove points further away than $r_{\text{pl}} = 20 \, \textup{m}$ as they are noisier than short-range points.
After reducing the point count, we call the stamped payload cloud $^{\mathtt{S}^t}\mathcal{P}_{\text{pl}}$.

\subsection{Terrain Model based on Cloth Simulation Filtering} 
For modeling sloping forest terrain, we generate a local DTM $^{\mathtt{S}^t}\mathcal{M}_\textup{DTM}^\text{local}$ for $^{\mathtt{S}^t}\mathcal{P}_{\text{pl}}$ using the cloth simulation filter proposed by Zhang \etalcite{zhang2016rs}.
We associate it with the SLAM pose graph so that we can combine the set of $^{\mathtt{S}^t}\mathcal{M}_\textup{DTM}^\text{local}$ into a global DTM $^\mathtt{M}\mathcal{M}_\textup{DTM}^\text{global}$ to extract tree traits (\secref{sec:spatio_temporal_aggregation}).

\subsection{Voronoi-Inspired Tree Segmentation}
To cluster the trees, we propose an adaptation to the algorithm introduced by Cabo \etalcite{cabo2018ijaeog}, where the authors clustered TLS point clouds. Cabo et al. normalized the floor height of the point cloud and cropped it between heights where they expected no foliage. After clustering the cropped sections, they fit tree axes to the clusters using principal component analysis, and ultimately assigned points to tree instances by distance to the closest tree axis following the Voronoi paradigm. 

We extend their approach by introducing non-maximum suppression (NMS), where we fit tree axes to three cropped sections---instead of one---and choose the best fit using a fitness function.
After normalizing the floor height of $^{\mathtt{S}^t}\mathcal{P}_{\text{pl}}$ using $^{\mathtt{S}^t}\mathcal{M}_\textup{DTM}^\text{local}$, we align the cloud with gravity by transforming it into the map frame $\mathtt{M}$.
Now, we crop the cloud at three height intervals and cluster the crops using Density-Based Spatial Clustering of Applications with Noise (DBSCAN) \cite{ester1996kdd} resulting in cluster point clouds $\{\mathcal{P}_\text{cluster}^{k^t}\}_{k^t}$. 
We fit a cylinder to each cluster by estimating two circles (\secref{sec:tree_reconstruction}) fitted to slices on the top and bottom of the cluster.
After this, we use the fitness function $\phi_{k^t}$ described in \eqref{eq:fitness_function}, employing the distance $d(p_i, a_{k^t})$ from point $p_i \in \mathcal{P}_\text{cluster}^{k^t}$ to axis $a_{k^t}$, the cylinder's radius $r_{k^t}$, and the Heaviside function $H(x)$.
This function penalizes points inside the cylinder and prefers points close to its surface.

Using $\phi_{k^t}$, we apply NMS to select the best cylinders, whose main axes build the final set of axes $\{a_\text{NMS}^l\}_l$.

\begin{equation}
  \label{eq:fitness_function}
  \phi_{k^t} = \frac{N_{1.2*r_{k^t}}}{{N_{0.5*r_{k^t}}}} \;\;\;\; N_{\theta} = \sum_i H(d(p_i, a_{k^t}) - \theta)
\end{equation}

Finally, we compute distances from every point in the height-normalized point cloud to all $a_\text{NMS}^l$.
After undoing the height-normalization and transforming the point clouds back into $\mathtt{S}^t$, we arrive at cluster point clouds $^{\mathtt{S}^t}\mathcal{P}_{\text{cluster}, l}$.

\subsection{Spatio-Temporal Aggregation}
\label{sec:spatio_temporal_aggregation}
In order to keep the map globally consistent over time, the tree manager uses $^\mathtt{M}\mathbf{T}_{\mathtt{S}^t}$ from the SLAM pose graph to transform raw measurements, which are stored in the sensor frame $\mathtt{S}^t$, into map frame $\mathtt{M}$.
Whenever loop closures occur, the SLAM system updates $^\mathtt{M}\mathbf{T}_{\mathtt{S}^t}$, which keeps the pose graph and thereby our reconstructions globally consistent.

To build $^\mathtt{M}\mathcal{M}_\textup{DTM}^\text{global}$, which is used to locate the measurement height for the DBH, all local DTMs are converted into map frame $\mathtt{M}$.
For smooth blending of the local models, we generate weights for every vertex of the DTMs using the function described in \eqref{eq:confidence_weights}.
It enables a C0-continuous transition between local DTMs using their width $w_\textup{DTM}$ and length $l_\textup{DTM}$ as well as the sensor position $\bm{x}_\text{sensor}$.
Using rays sampled on a regular grid and an efficient ray-to-mesh intersection algorithm by Wald \etalcite{wald2014acmgraphics}, we build $^\mathtt{M}\mathcal{M}_\textup{DTM}^\text{global}$ by computing the weighted average of the local DTM heights for every ray.

\begin{equation}
\label{eq:confidence_weights}
    w(\bm{x}) = 1 - \frac{||\bm{x} - \bm{x}_\text{sensor}||_2}{\min(l_\text{DTM}, w_\text{DTM}) / 2}
\end{equation}

In addition to $^\mathtt{M}\mathcal{M}_\textup{DTM}^\text{global}$, the Tree Manager also maintains a database of tree instances.
Whenever a clustering result becomes available, every cluster $^{\mathtt{S}^t}\mathcal{P}_{\text{cluster}, l}$ is compared to the current database of trees.
$^{\mathtt{S}^t}\mathcal{P}_{\text{cluster}, l}$ is either added to an existing tree instance if it is close, or a new instance is created.
Using the timestamps of the clusters for association, the Tree Manager regularly realigns all clusters in all trees using the most recent $^\mathtt{M}\mathbf{T}_{\mathtt{S}^t}$ from the SLAM pose graph.

To make the best use of the available compute resources, we require certain coverage conditions on every tree before it is reconstructed.
The first condition is a maximum distance of the sensor from the tree of at least $d^{\text{reco}}_{\text{min}}$, which ensures that the LiDAR with its limited field of view has scanned points sufficiently high up the tree.
The second condition is the coverage angle $\alpha^{\text{reco}}_{\text{min}}$ making sure that there are sufficient measurements from around the tree. 
In our experiments values of $d^{\text{reco}}_{\text{min}} = 10 m$ and $\alpha^{\text{reco}}_{\text{min}} = \pi$ gave reasonable results.

\subsection{Tree Reconstruction and Tree Trait Extraction}
\label{sec:tree_reconstruction}
Once the reconstruction criteria of the Tree Manager are fulfilled, we reconstruct the tree as a stack of oblique cone frustums between circles at regular heights (\figref{fig:stacked_frustums}).
To generate a circle modeling the tree's crosssection at a certain height, we slice each cluster in a tree instance resulting in 2D point clouds.
The biggest challenges here are to reject outliers originating from twigs and branches and to be robust against noise introduced by the sensor and the odometry while retaining real-time performance.

De Conto \etalcite{deconto2017cea} proposed an effective algorithm to achieve this robustness by first filtering outliers using the Hough Transform \cite{hough1962patent} applied to circles \cite{duda1972use} and then to fit a circle in the least squares sense to the remaining points.
Using the Hough Transform, one can detect circular shapes in a bitmap of edges by using circles of variable centers and radii to vote within a discretized Hough space.
For circles, the Hough space is three-dimensional with two coordinates for the circle center and one for the radius. 

To apply the Hough Transform to our point cloud setting, one has to rasterize the points by aggregating them into a bitmap.
In our experiments, desirable results required fine rasterization, which incurred high memory and computational costs.
To mitigate this, we opted to use the sampling-based Randomized Hough Transform (RHT) \cite{ballard1981pr, xu1990prl}, which Jiang \etalcite{jiang2012optik} have demonstrated to be very efficient for fitting circles.

\begin{figure}
  \centering
  \begin{minipage}{.8\linewidth}
    \centering
    \includegraphics[height=1.5in]{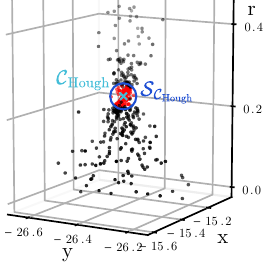}
    \caption{Point cloud $\mathcal{P}_\text{Hough}$ of triplet-circles represented in the Hough space. At a point $\mathcal{C}_\text{Hough}$ with a sphere $\mathcal{S}_{\mathcal{C}_\text{Hough}}$ around it containing many triplet-circles, a good circle fit can be found.}
    \label{fig:pointcloud_houghspace}
  \end{minipage} 
\end{figure}

To implement RHT, we sample triplets of points and explicitly fit circles to them, which we call triplet-circles. 
Inspired by \cite{jiang2012optik}, we employ density-weighted sampling where points with close neighbors are more likely to be part of a triplet.
We transform all triplet-circles into the Hough space where they form a set of votes $\mathcal{P}_\text{Hough}$ following the Hough paradigm.
The optimal circle fit is then found by locating the point $\mathcal{C}_\text{Hough}$ in the Hough space, where the most votes are concentrated.
To find this point, we represent $\mathcal{P}_\text{Hough}$ in an octree which allows us to efficiently find the point in $\mathcal{P}_\text{Hough}$ with the most neighbors within a sphere $\mathcal{S}_{\mathcal{C}_\text{Hough}}$ of fixed radius around it (\figref{fig:pointcloud_houghspace}).
We demonstrate the benefit of using RHT over different outlier rejection mechanisms in \secref{sec:hough_ablations}.

To optimize computational efficiency, we constrain the Hough space to consider only circles close to the previous circle in the x-y plane of $\mathtt{M}$.
For this to work reliably, we need a robust initialization, which we achieve by using a non-maximum suppression (NMS) on the first three reconstructions.

To combine circle estimates from the individual clusters into a single estimate of the trunk's cross-section, we follow the approach of Hyypp\"a \etalcite{hyyppa2020jprs} by translating the sliced tree clusters such that the centers of the Hough-circles align.
After alignment, we merge the sliced clusters into a single point cloud to which we fit the final circle in the least-squares sense, for which Bullock \etalcite{bullock2006dtc} proposed an explicit algorithm.

For visualization purposes, we also generate canopy meshes.
These can be efficiently computed by fitting a convex hull to all canopy points.
We treat points more than \SI{2}{\meter} away from $^\mathtt{M}\mathcal{M}_\textup{DTM}^\text{global}$ and greater than twice the diameter away from the stem axis as canopy points.

Once reconstructions of terrain and trees are available, we visualize them in real-time, which enables assessment of the scanning quality and the coverage of the plot.
Immediately after the mapping session ends, we export the results into an industry-standard representation for a marteloscope.

\section{Experimental Evaluation}
\label{sec:exp}

The focus of this work is the implementation of a real-time pipeline that reconstructs individual trees and extracts their tree traits.
We present our experiments to show the capabilities of our method and to support our key claims, which are:
(i) The extraction of relevant tree traits with accuracy competitive to state-of-the-art approaches,
(ii) the generation of a globally consistent map of the forest,
(iii) the robust fitting of oblique cone frustums to the trees in the presence of heavy sensor noise
(iv) the ability to run on a mobile system.

\setlength{\tabcolsep}{5pt}
\renewcommand\cellset{\renewcommand\arraystretch{0.5}}
\begin{table}[]
  \centering
  \begin{tabular}{cc|ccc|c}
  \multicolumn{1}{l|}{}                                           & Plot                           & Conifer & Mixed   & Deciduous & All    \\ \hline\hline
  \multicolumn{2}{c|}{Detection Recall $\uparrow$}                                              & 98.3\%  & 98.6\%  & 97.4\%    & 98.1\% \\\hline
  \multicolumn{1}{l|}{\multirow{3}{*}{\makecell{DBH}}}            & RMSE {[}cm{]} $\downarrow$     & 1.18    & 2.22    & 2.38      & 1.93   \\
  \multicolumn{1}{l|}{}                                           & Bias {[}cm{]} $\downarrow$     & 0.02    & 0.34    & 2.05      & 0.65   \\
  \multicolumn{1}{l|}{}                                           & Std {[}cm{]} $\downarrow$      & 1.17    & 2.18    & 1.04      & 1.81   \\ \hline
  \multicolumn{1}{l|}{\multirow{2}{*}{\makecell{RMSE \\ Stem}}}   & Diameter {[}cm{]} $\downarrow$ & 2.91    & 3.14    & 3.12      & 3.02   \\
  \multicolumn{1}{l|}{}                                           & Center {[}cm{]} $\downarrow$   & 5.78    & 9.56    & 14.88     & 8.05   \\ \hline
  \multicolumn{1}{l|}{\multirow{2}{*}{\makecell{Mean \\ Height}}} & Ours {[}m{]} $\uparrow$        & 8.36    & 6.30    & 3.30      & 6.12   \\
  \multicolumn{1}{l|}{}                                           & TLS {[}m{]} $\uparrow$         & 17.16   & 15.22   & 5.94      & 10.22   
  \end{tabular}
  \caption{Evaluation of our pipeline on three different plots. We report the RMSE of DBH estimates (relative to manual measurements) and the RMSE of the stem diameter and curvature measured along the entire stem (relative to a TLS based model). Additionally, we measured the mean height of the reconstructed stems and the detection recall of the clustering algorithm.}
  \label{tab:accuracy}
\end{table} 

\subsection{Tree Trait Estimation Accuracy}
The first experiment evaluates the quality of our reconstructions and demonstrates that we can estimate tree traits with accuracy competitive with the state of the art.
We considered three plots located in a forest in Stein am Rhein, Switzerland consisting of coniferous (58 trees), broad-leaf (163 trees) and a mixture of the two (70 trees). 
We expect conifer trees with large diameters and a sparse under-canopy to be easier to reconstruct than broad-leaf trees with more vegetation close to the ground as well as a complex branching structure. 

The campaign was executed in March 2023 which involved MLS scanning with our Mobile Mapping System and manual measurements of the DBH by a team of professional foresters. We use these manual measurements as a ground truth to compare our \emph{DBH estimates} to.
We also want to assess the location of the stem's center as well as the diameter of the tree trunk along the entire height, which we call \emph{stem curve}.
For this assessment, we need more descriptive measurements, which we obtain by building tree models from TLS scans with a resolution of \SI{1}{\centi\meter}.
For this experiment, we evaluated our system by simulating online data acquisition and tree reconstruction after data acquisition has ended.

\subsubsection{Detection Recall}
We started by evaluating our clustering algorithm to determine if we were able to detect all the trees present using the MLS data.
As reported in \tabref{tab:accuracy}, our method was able to detect 98.08 \% of all trees.

\subsubsection{DBH Estimation}
\begin{figure}
  \centering
  \begin{minipage}{0.4\linewidth}
    \centering
    \includegraphics[width=\linewidth]{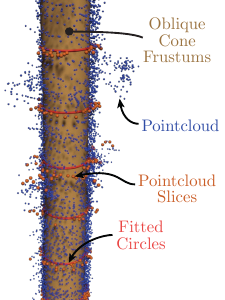}
    \caption{Stem reconstruction as a stack of oblique cone frustums.}
    \label{fig:stacked_frustums}
  \end{minipage}\hfill
  \begin{minipage}{0.5\linewidth}
    \centering
    \includegraphics[width=\linewidth]{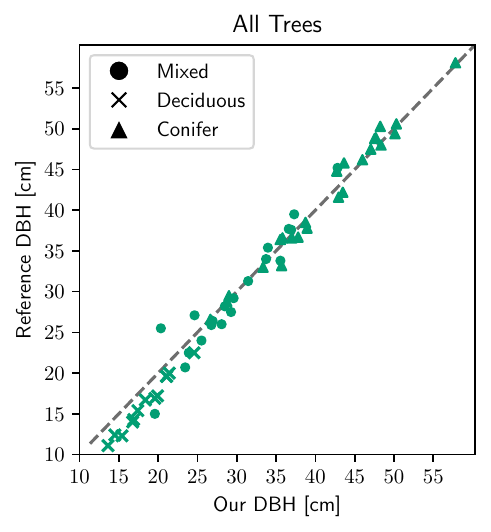}
    \caption{Scatter plots of the DBH estimates of our reconstructions against reference measurements.}
    \label{fig:scatter_dbh}
  \end{minipage}
\end{figure}

Next, we assessed the accuracy of our DBH estimates. For ground truth, we used the manual reference measurements of the tree diameters conducted by the foresters at a height of \SI{1.3}{\meter} above the ground, which we associated with the MLS scans using an AprilTag-based matching system.
We report the results in \tabref{tab:accuracy} and visualize the results in \figref{fig:scatter_dbh}.
As expected, \tabref{tab:accuracy} shows that estimates for conifer trees are the most accurate with an RMSE of \SI{1.18}{\centi\meter}, while the broad-leaf trees are the most challenging with an RMSE of \SI{2.38}{\centi\meter}.
The mixed plot is in between with an RMSE of \SI{2.22}{\centi\meter}.

\subsubsection{Stem Curve Estimation}
Additionally, we evaluated the accuracy of our estimates of the entire stem curve by comparing our circle centers and diameters along the stem to tree models based on the TLS measurements.
To build these models, we sliced the TLS tree clouds at regular intervals, annotated stem points and fitted circles to them in the least-squares sense.
In \tabref{tab:accuracy}, we report an RMSE of diameter estimates along the stem of \SI{3.02}{\centi\meter} and an RMSE of the stem curvature of \SI{8.05}{\centi\meter}---averaged over all plots.

\subsubsection{Tree Height}
Finally, we compared the heights of the reconstructed stems. 
For the TLS measurements, the stem cross-section could no longer be reliably estimated above heights of \SI{10.22}{\meter}, our reconstructions could only estimate up to an average height of \SI{6.12}{\meter} (\tabref{tab:accuracy}).
We attribute this to the sensor configuration, which has a limited field of view and thus cannot measure high up the tree.

\subsection{Global Consistency}
We designed the second experiment to demonstrate the importance of having a SLAM system ensuring globally consistent maps. 
For that, we let the pipeline run on the conifer plot again, but this time we disabled the trajectory updates in the Tree Manager.
We did not disable the loop closures for the SLAM system, as drift accumulated over the entire trajectory would have made tree associations impossible. 
This implies that this experiment only considers the effect of the odometry drift in between loop closures. 

\begin{figure}
  \centering
  \begin{minipage}{.5\linewidth} 
    \centering
    \includegraphics[trim = 15 10 35 5, clip, width=\linewidth]{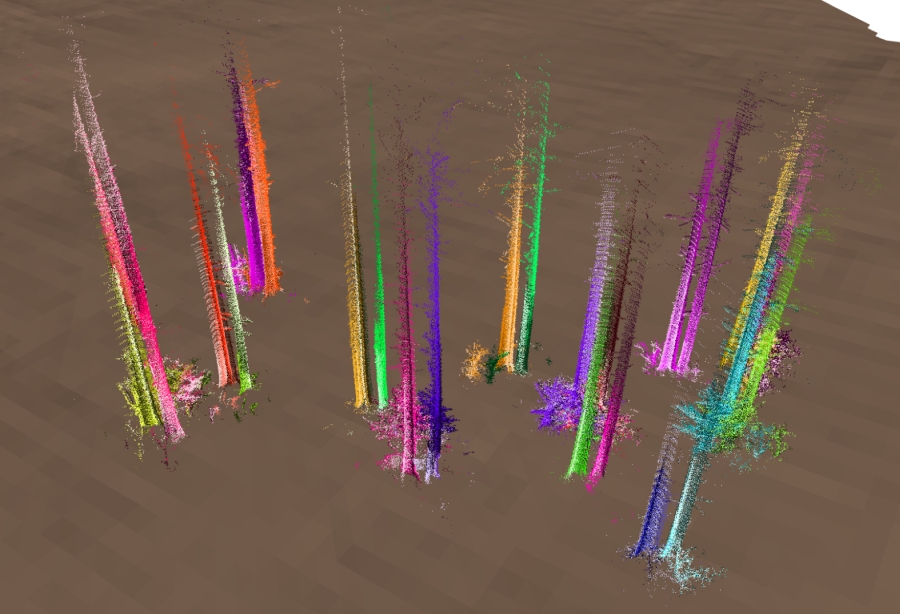}
    \caption{Incorrect tree association prior to a loop closure. Unique trees are represented as two different instances (different hues).}
    \label{fig:doubled_instances}
  \end{minipage}\hfill
  \begin{minipage}{.45\linewidth}
    \centering
    \includegraphics[width=0.95\linewidth]{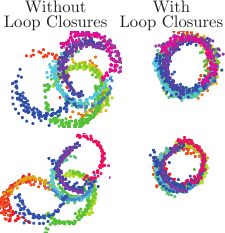} %
    \caption{Misaligned cluster point clouds due to odometry drift. Loop closure corrections are able to realign them.} 
    \label{fig:double_walls}
  \end{minipage}
\end{figure}

Visually, the misalignment of the point clouds is shown in \figref{fig:double_walls}, but because of the realignment procedure described in \secref{sec:tree_reconstruction}, our pipeline could still reconstruct stems.

A second effect is the failure of our pipeline to reliably merge tree clusters into a single tree instance (\figref{fig:doubled_instances}). This is due to the odometry drift being larger than our distance threshold for merging.

Ultimately, disabling loop closures in the tree manager should cause a reduction in the quality of the stem curve, which becomes apparent in \tabref{tab:global_consistency}.
We report a significant decrease of the RMSE in fitting the stem curve, which is to be expected if the point clouds are severely misaligned.

\begin{table}
  \centering
  \begin{tabular}{c|c c c}
    RMSE [cm] & DBH & Stem Curve & Stem Diameter \\
    \hline\hline\rule{0pt}{1.8ex}
    w/o loop closures & 5.85 & 35.61& 3.76 \\
    w/ loop closures & 1.18 & 4.65 & 2.82 \\
  \end{tabular}
  \caption{Comparison of running the pipeline with or without loop closures. Evaluated is the RMSE in the fitting of DBH, stem curve and stem diameter. All results are averaged over three runs.}
  \label{tab:global_consistency}
\end{table}

\subsection{Ablations}
We designed a third set of experiments to demonstrate the benefit of other central components of our pipeline and their impact on reconstruction quality.

\subsubsection{Randomized Hough Transform}
\label{sec:hough_ablations} 
To demonstrate the advantage of the Randomized Hough Transform (RHT) algorithm, we compared it with the classical Hough algorithm, regular RANSAC fitting \cite{fischler1981acm}, and RANSAC*, where we applied the density-weighted subsampling procedure as described in \secref{sec:tree_reconstruction}.
For RANSAC and RHT we used 500 algorithm iterations.

In \tabref{tab:ransahc_ablations}, we report the RMSE of the DBH estimates for all ablations averaged over three runs on all three datasets. RHT outperforms the alternatives in terms of RMSE.
Regarding timing, the classical Hough algorithm is faster, which we attribute to a less expensive mechanism for vote aggregation in the Hough space.

\begin{figure}
  \centering
  \begin{minipage}{0.32\linewidth}
    \centering
    \scriptsize{Hough}
  \end{minipage}
  \begin{minipage}{0.32\linewidth}
    \centering
    \scriptsize{RANSAC}
  \end{minipage}
  \begin{minipage}{0.32\linewidth}
    \centering
    \scriptsize{RHT}
  \end{minipage}
  \begin{minipage}{0.32\linewidth}
    \centering
    \includegraphics[width=\linewidth]{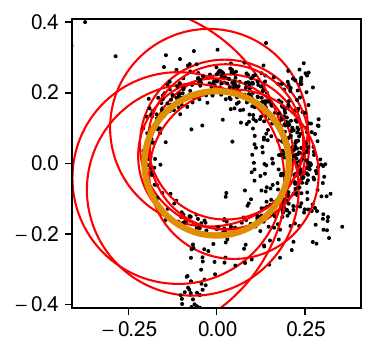}
  \end{minipage}
  \begin{minipage}{0.32\linewidth}
    \centering
    \includegraphics[width=\linewidth]{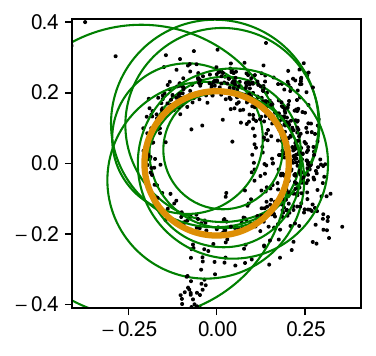} 
  \end{minipage}
  \begin{minipage}{0.32\linewidth}
    \centering
    \includegraphics[width=\linewidth]{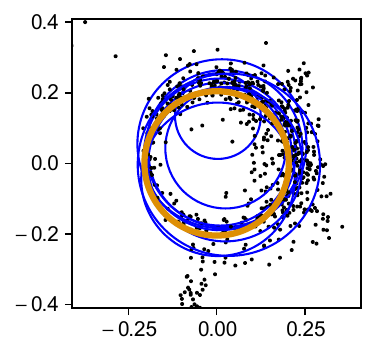}
  \end{minipage}
  \begin{minipage}{0.32\linewidth}
    \centering
    \includegraphics[width=\linewidth]{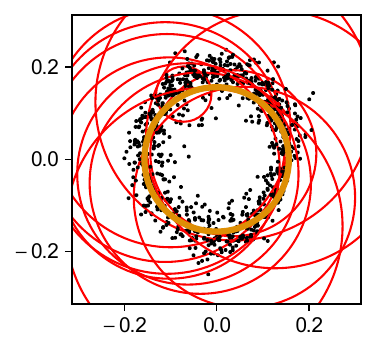}
  \end{minipage}
  \begin{minipage}{0.32\linewidth}
    \centering
    \includegraphics[width=\linewidth]{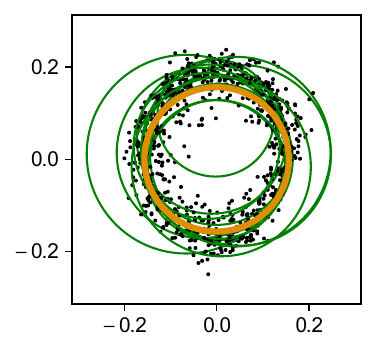} 
  \end{minipage}
  \begin{minipage}{0.32\linewidth}
    \centering
    \includegraphics[width=\linewidth]{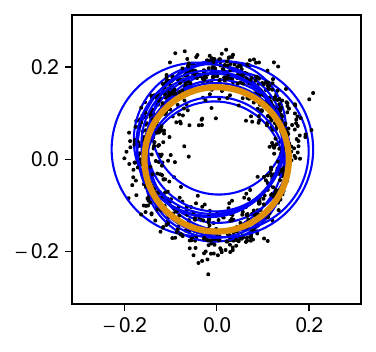}
  \end{minipage}
  \caption{Examples of circle estimates on two point clouds with noise. The Ransac and Hough circle fit overestimates the diameter. RHT is able to fit the circle more accurately. The TLS circle fit is shown in orange. Note that the black points represent points from different clusters as in \figref{fig:double_walls}}.
  \label{fig:ransahc_ablations}
\end{figure}

In \figref{fig:ransahc_ablations}, we present a qualitative analysis to give an intuition for why regular RANSAC and the Hough algorithm are inferior.
Firstly, the Hough algorithm tends to overestimate the diameter of a cluster.
Secondly, regular RANSAC struggles in the presence of branches or other sources of noise.
We attribute both of these phenomena to the low sampling density of our point clouds, which increases the relative impact of noise.
This biases the inlier counting mechanism of Hough and RANSAC towards larger circles.
Meanwhile, RHT with a more robust mechanism for inlier detection, does not suffer from this bias.

\begin{table}
  \centering
  \begin{tabular}{c|c c c|c|c}
    \multirow{2}{*}{Algorithm} & \multicolumn{4}{c|}{RMSE [cm]} & \multirow{2}{*}{\makecell{Time \\ {[}ms{]}}} \\
    \cline{2-5}
    & Conifer & Deciduous & Mixed & All & \\
    \hline\hline\rule{0pt}{1.8ex}
    Hough & 8.43 & 12.45 & 5.22 & 7.68 & \textbf{115}\\
    RANSAC & 5.81 & 3.48 & 2.69 & 4.27 & 247\\
    RANSAC* & 2.78 & 3.88 & 2.67 & 3.26 & 275\\
    RHT & \textbf{1.18} & \textbf{2.38} & \textbf{2.22} & \textbf{1.93} & 191\\
  \end{tabular}
  \caption{Ablation study for different variants of robust circle fitting. We report the RMSE of the DBH estimates for the different versions averaged over three runs across all three of our datasets. We also report the timing of the algorithms.}
  \label{tab:ransahc_ablations}
\end{table}

\subsubsection{Coverage Angle} 
\begin{figure}
  \centering
  \includegraphics[width=0.9\linewidth]{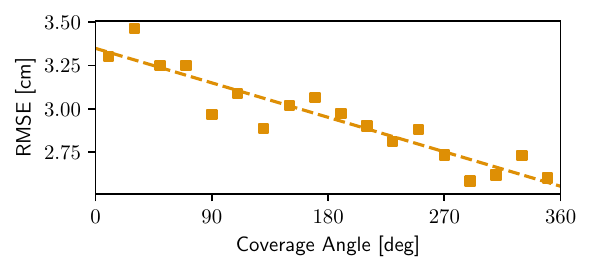} 
  \caption{Accuracy of the DBH estimates as a function of coverage angle. As more of the tree is observed the DBH RMSE decreases.}
  \label{fig:ablation_coverage}
\end{figure}

This experiment was conducted to support our assumption that larger coverage angles, i.e. the range of directions the tree is scanned from,  are beneficial for reconstruction quality.
We reconstructed every tree of the three plots several times, each time removing clusters and noting the reduced coverage angle. 
We grouped results in buckets of $20^\circ$ increments and for each bucket computed the RMSE of diameter estimates along the stem curve with respect to the TLS dataset.
We note a clear trend where increasing the coverage angle decreases the error of the estimate (\figref{fig:ablation_coverage}).

This suggests that moving the sensor through the plot and measuring the trees from different angles is beneficial for accurate stem reconstruction.

\subsection{Performance}

This experiment explores the computation time of the real-time pipeline.
We evaluated the runtime on a Simply NUC Topaz 3 featuring a Core i7-1165G7 with 4 cores, a base frequency of 2.8 GHz, and \SI{32}{\giga\byte} of RAM.
Payload clouds are accumulated over \SI{20}{\meter}, which on average took \SI{19.8}{\second}. 

\begin{figure}
  \centering
  \includegraphics[width=\linewidth]{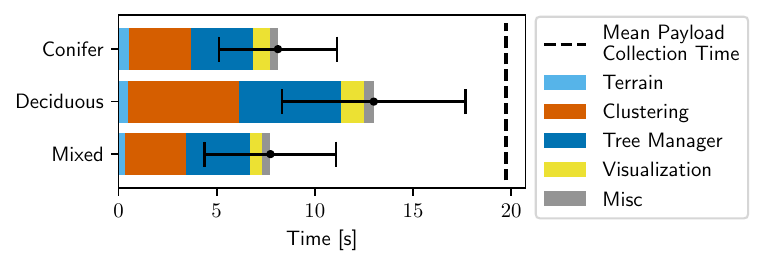} 
  \caption{Runtime of individual components of our pipeline given by mean and standard deviation. 
  With a time budget of \SI{19.8}{\second} per payload, our approach runs almost twice as fast as required.}
  \label{fig:runtime}
\end{figure}

\figref{fig:runtime} presents the average runtime of the components of our pipeline for our different datasets.
With an overall mean runtime of \SI{9.69}{\second} and a standard deviation of \SI{4.41}{\second}, the algorithm is able to run approximately twice as fast as the capture frequency.
Note that processing the deciduous plot consumes more time than the other two, which is due to the higher tree stem density.
Across all datasets, the computationally most expensive component is the Tree Manager, which handles the reconstructions.

We also report system memory usage (\tabref{tab:memory}) which, with an average of \SI{5.69}{\giga\byte}, is well within the capabilities of our system.
Perhaps the biggest advantage of our approach is in storage space. The average storage requirements of the final output data are in the range of \SI{160}{\mega\byte} for \SI{0.7}{hectare} including the point clouds of every tree, and only \SI{6}{\mega\byte} when storing just the reconstruction results. 
This compares to \SI{13.5}{\giga\byte} for the raw data and hundreds of gigabytes for a traditional TLS system, which makes our system three orders of magnitude more efficient in terms of storage.
This detail is critical as data acquisition is arduous and done in remote locations.
Furthermore, compared to the intrinsic value of the raw measurements, the cost of storage is immense.

\subsection{Real-time Demonstration on ANYmal Robot}
\begin{figure}
  \centering
  \includegraphics[width=\linewidth]{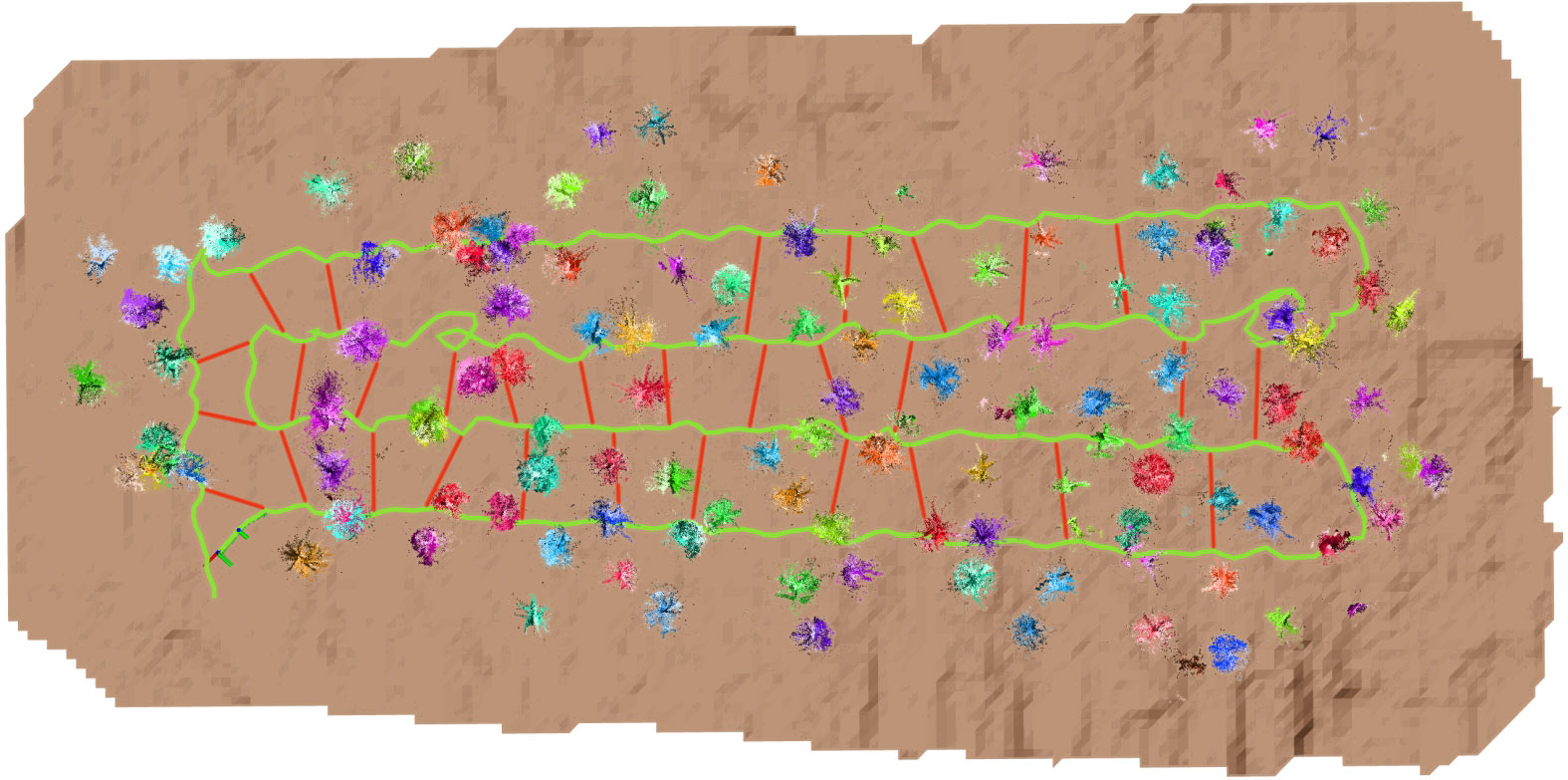} 
  \caption{Fully autonomous mission through a plot of the Forest of Dean with the ANYmal quadruped autonomously navigating the forest following a lawnmower pattern. On board the Mobile Robot, our pipeline created an online forest inventory which was visualized for the operators in real-time. Note the pose graph (green) being corrected using loop closures (red). }
  \label{fig:autonomous_mapping}
\end{figure}

In order to demonstrate this capability running live on a robot, we carried out a field trial in the Forest of Dean, UK.
We used the ANYmal quadruped robot to autonomously carry the Mobile Mapping System through the forest with all reconstructions carried out in real-time and visualized online. This demonstration is presented in \figref{fig:autonomous_mapping} and the supplementary video.
(A complete description of the robot autonomy system is outside the scope of this paper.)

\begin{table}[]
  \centering
  \begin{tabular}{cc|ccc}
                                                &            & Conifer                & Deciduous              & Mixed                  \\ \hline\hline
  \multicolumn{2}{c|}{Memory}                                & \SI{5.18}{\giga\byte}  & \SI{6.75}{\giga\byte}  & \SI{5.15}{\giga\byte}  \\ \hline
  \multicolumn{1}{c|}{\multirow{2}{*}{Storage}} & w/ clouds  & \SI{158.0}{\mega\byte} & \SI{187.7}{\mega\byte} & \SI{143.1}{\mega\byte} \\ \cline{2-2}
  \multicolumn{1}{c|}{}                         & w/o clouds & \SI{1.5}{\mega\byte}   & \SI{9.2}{\mega\byte}   & \SI{8.0}{\mega\byte}  
  \end{tabular}
  \caption{Memory (during runtime) and Storage requirements (afterwards) for final output data (with and without storing the point clouds for every tree).}
  \label{tab:memory}
\end{table}

\section{Conclusion}
\label{sec:conclusion}
In this paper, we presented a novel approach for real-time tree inventory and reconstruction in forest environments.
Our method incrementally reconstructs hectare-sized forest plots by accumulating a local submap (a payload) and clustering and reconstructing tree instances within it.
Payloads are incrementally merged and global consistency of the full map is maintained by leveraging a pose graph SLAM system.

Compared to other state-of-the-art methods that compute the result in post-processing, our approach produces a dataset of tree instances and tree traits (such as the DBH and stem curve) during data acquisition.
Additionally, we can report the state of the reconstruction in real-time, helping the forester---or an autonomous robot---to adjust the mission to e.g. increase coverage for better reconstructions.

We evaluated our pipeline on three forest datasets with different compositions and showed that we can estimate tree traits with an accuracy competitive with the state of the art.
Averaging across deciduous and conifer plantations, our method achieved an RMSE for DBH estimates of \SI{1.93}{\centi\meter}.

Focusing on conifer plantations, our accuracy was measured as \SI{1.18}{\centi\meter} while a state-of-the-art method from Hyypp\"a et al. achieved an RMSE of \SI{0.6}{\centi\meter} \cite{hyyppa2020jprs_drones}.
We note that our approach due to its realtime nature required significantly less computation time as well as an order of magnitude cheaper hardware.

Using our three datasets we supported all claims made in this paper.
The experiments suggest that online estimation of tree traits is feasible and can deliver faithful reconstructions of trees.
For further development, our pipeline is a good starting point to use other sensor modalities to estimate more tree traits such as the tree species from RGB images.
Furthermore, the biggest drawback of little scanning density for high-up regions of the tree could be addressed by specifically adding a LiDAR sensor for scanning deeper into the canopy.

Considering the light weight of the pipeline as well as the Mobile Mapping System itself, we believe that with improvements in remote sensing hardware, building online forest inventories can be a valuable tool to quickly generate accurate models of forests. 

\balance
\bibliographystyle{plain_abbrv}
\section*{Acknowledgement}
This work has been funded by the Horizon Europe project DigiForest(101070405) and a Royal Society University Research Fellowship (M. Fallon).
We acknowledge the assistance of the Swiss Federal Institute for Forest, Snow and Landscape Research (WSL) who carried out the manual measurements at Stein am Rhein as well as by PreFor who collected the TLS dataset.

\bibliography{glorified,new}

\end{document}